\documentclass[lettersize,journal]{IEEEtran}
\usepackage{amsmath,amsfonts}
\usepackage{algorithmic}
\usepackage{algorithm}
\usepackage{array}
\usepackage[caption=false,font=normalsize,labelfont=sf,textfont=sf]{subfig}
\usepackage{textcomp}
\usepackage{stfloats}
\usepackage{url}
\usepackage{verbatim}
\usepackage{graphicx}
\usepackage{booktabs}
\usepackage{multirow}
\usepackage{multicol}
\usepackage{xcolor}
\usepackage{colortbl}
\usepackage{cite}
\usepackage{pifont}
\usepackage{hyperref}  
\usepackage{cleveref}  
\usepackage{amssymb} 
\usepackage{textcomp}    
\usepackage{fontawesome} 
\usepackage{amsmath}
\usepackage{chngcntr}
\counterwithout{figure}{section}

\begin{document}
\bstctlcite{IEEEexample:BSTcontrol}

    \title{DREAM: Extending Vision-Language Models with Dual-Objective Encoding for Cross-Modal Retrieval }
    
 \author{Kaleem Ullah, Altaf Hussain, Muhammad Munsif, Sung Wook Baik* 
  
\thanks{This work was supported by National Research Foundation of Korea(NRF) grant funded by the Korea government (MSIT), Grant /AwardNumber:(RS-2023-NR076686))
\indent Kaleem Ullah and Sung Wook Baik are with Sejong University, Seoul 143-747, South Korea, Altaf Hussain is with KAIST InnoCORE PRISM-AI Center, Korea Advanced Institute of Science and Technology (KAIST), Daejeon, 34141, Republic of Korea and Muhammad Munsif is with Ulsan National Institute of Science and Technology (UNIST), Ulsan, 44919, Republic of Korea. (Email: \href{mailto:}{kaleemullah@sju.ac.kr}, \href{mailto:}{a.hussain@ieee.org}, \href{mailto:}{munsif@ieee.org},
\href{mailto:}{sbaik@sejong.ac.kr}, (*Corresponding author: Sung Wook Baik)

  }
  }

\maketitle

\maketitle

\begin{abstract}
 In today’s media-driven world, the exponential growth of video content across domains such as surveillance, education, and entertainment has made retrieving semantically relevant videos via natural language queries increasingly critical. Early video retrieval systems relied on handcrafted features or shallow cross-modal mappings, limiting their ability to capture complex semantics and temporal dynamics. While large-scale vision-language models (VLMs) have improved cross-modal alignment, challenges remain in modeling fine-grained temporal dependencies and nuanced linguistic structures. In this paper, we introduce DREAM: Dual-path Representation Enhancement and Alignment Model, a novel multimodal framework that addresses these limitations through enhanced visual and textual encoding. DREAM incorporates a hybrid language modeling strategy that combines masked and permuted language modeling objectives to capture both local and global linguistic semantics. On the visual side, we design a hierarchical vision encoder with cascaded group attention, which integrates spatial and temporal information through multi-stage token interaction and coarse-to-fine attention refinement. \textcolor{black}{We validate DREAM through comprehensive evaluations on the widely-used MSRVTT, MSVD and LSMDC benchmark datasets, where it achieves new state-of-the-art R@1 scores of 49.4\%, 49.7\% and 27.3\%, respectively.} Qualitative analyses further show the model’s ability to maintain coherent attention across frames and align complex queries with dynamic video content. These findings underscore the effectiveness of hierarchical attention and dual-objective textual modeling in enabling robust, context-aware video retrieval, and pave the way for future research in advancing cross-modal representation learning.
\end{abstract}

\begin{IEEEkeywords}
Text-Video Retrieval, Multimodal Understanding, Cross-Modal Alignment, Video Understanding, Vision-Language Models.
\end{IEEEkeywords}
\section{Introduction}
\IEEEPARstart{V}{ideo} has become the dominant form of media in today’s multi-modal world, driven by the rapid expansion of online platforms and the widespread deployment of cameras across domains such as surveillance, entertainment, education, and autonomous driving\cite{tang2025video}. As a rich and immersive medium, video offers greater expressiveness and user engagement than static images or text-based content. Unlike still images or textual descriptions, videos capture complex, evolving activities and richer semantics over time.

\textcolor{black}{This surge in video content, with millions of new videos produced daily has created a pressing need for automated systems capable of understanding, organizing, and retrieving relevant visual information. Consequently, video understanding technologies have seen significant progress, powering applications in classification, summarization, and retrieval \cite{qu2025mvp}\cite{ma2025multi}. Among these, cross-modal video-text retrieval has emerged as a critical research area, enabling systems to locate semantically relevant videos in response to natural language queries.}

\textcolor{black}{Recent advances in text-video retrieval have been driven by transformer-based vision-language models and contrastive pretraining frameworks. Methods such as CLIP4Clip \cite{luo2022clip4clip}, MV-Adapter \cite{jin2024mv}, VILT-CLIP \cite{wang2024vilt}, and MUSE \cite{tang2025muse} introduced stronger spatiotemporal reasoning through global attention, multiscale processing, or lightweight temporal fusion. While these approaches improve video representation quality, many rely on flat or uniform attention mechanisms that lack hierarchical refinement and fail to capture fine-grained temporal dynamics in complex scenes. On the language side, models such as BLIP \cite{li2022blip}, BridgeTower \cite{xu2023bridgetower}, Long-CLIP \cite{zhang2024long}, and TeachText \cite{croitoru2021teachtext} have demonstrated the benefits of generative supervision or distillation for enhancing semantic expressiveness. However, most retrieval systems still rely on a single modeling paradigm either masked or contrastive and thus fail to jointly capture local token dependencies and global sentence structure. These limitations highlight the need for retrieval frameworks that integrate richer linguistic modeling with hierarchical and discriminative spatiotemporal representation learning.}

To address these challenges, we propose a unified retrieval framework DREAM, that jointly models visual and textual representations using a hybrid language encoder and a hierarchical vision encoder. On the text side, we introduce a hybrid language modeling strategy that combines masked and permuted objectives to capture both local context and global semantic structure. We adopt this hybrid strategy to leverage the complementary strengths of both objectives, capturing fine-grained dependencies as well as broader contextual flow. On the visual side, we develop a hierarchical vision encoder with cascaded group attention (CGAT) to refine spatiotemporal features through multi-stage token interaction. This hierarchical design effectively captures multi-scale spatial and temporal dependencies for more expressive visual representation. \textcolor{black}{Unlike CLIP-based methods, which rely on simple frame-level feature aggregation, our hierarchical design progressively refines spatial and temporal dependencies, capturing multi-scale spatiotemporal cues for more expressive visual representations. Together, these approaches enable more accurate and context-aware video retrieval, as demonstrated through our experiments on the benchmark datasets.} Furthermore, to provide a concise overview and impact of our approach, we summarize the key contributions of this work as follows.
\begin{figure*}
    \centering
    \includegraphics[width=\linewidth]{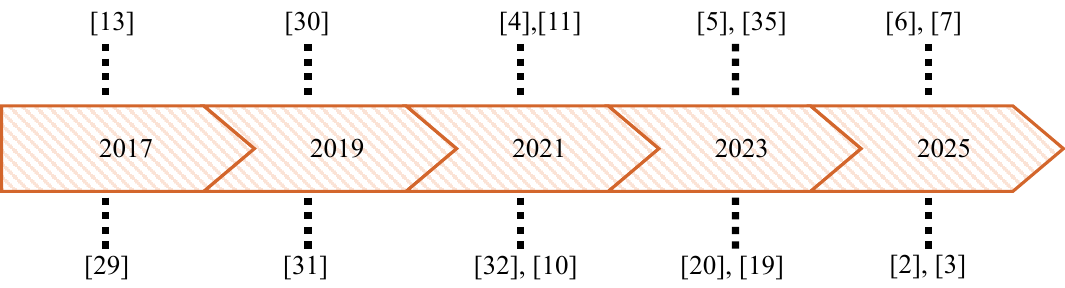}
    \caption{Progress of methodologies in text-to-video retrieval. With the establishment and broader adoption of standard benchmarks, research has shifted from earlier feature-engineering and statistical matching pipelines toward predominantly deep learning–based representation learning and video–text alignment methods.}
    \label{fig:literature}
\end{figure*}
\subsection{Main Contributions:}
 
\begin{itemize}
    \item We propose a novel dual-path text encoding framework that enhances semantic understanding through the integration of Masked Language Modeling (MLM) and Permuted Language Modeling (PLM) objectives. This hybrid  structure captures both fine-grained local context and global sentence structure in textual queries, enabling more accurate interpretation of ambiguous and context-dependent queries. On top of a CLIP text encoder, our hybrid approach significantly improves the robustness and accuracy of text representations, leading to superior alignment with complex and temporally dynamic video content.
    \item We propose a visual encoder that combines hierarchical feature extraction with efficient attention mechanisms for video representation learning. Drawing on principles from efficient transformer architectures, the model integrates multi-stage token embeddings and a Cascaded Group Attention module to progressively distill spatial semantics. This design improves the encoder’s ability to capture fine-grained visual patterns across multiple spatial and temporal scales, resulting in richer and more discriminative video embeddings that encapsulate context for text-to-video retrieval.
    \item Extensive experiments on the MSR-VTT, MSVD and LSMDC datasets demonstrate the effectiveness of our proposed framework. Our model achieves state-of-the-art retrieval performance, surpassing previous methods with an R@1 of 49.4 on MSR-VTT, 49.7 on MSVD, and 27.3 on LSMDC along with the lowest median rank (MdR) on these datasets. These results highlight the strength of our design in capturing fine-grained video-text semantics and ensuring robust cross-modal retrieval.
\end{itemize}

The rest of this paper is organized as follows: Section~II reviews prior work along two main dimensions relevant to text-video retrieval, namely text modeling for cross-modal representation learning and video modeling for efficient spatiotemporal understanding, and identifies the limitations that motivate our approach. Section III details the proposed DREAM retrieval framework for cross-modal alignment mechanism. Section IV presents the experimental results, comparisons along with qualitative analysis. Finally, Section V concludes the article with a summary of findings, current limitations and potential directions for future work.

\section{Related Work}
\textcolor{black}{
Multimodal retrieval research has advanced primarily along two complementary dimensions: the development of expressive text encoders to better interpret natural language queries, and the design of efficient video encoders capable of capturing complex spatiotemporal patterns. Fig.~\ref{fig:literature} shows this evolution outlining the chronological development of key retrieval methods, these approaches have shaped the evolution of cross-modal retrieval from early image-based systems to modern transformer-driven frameworks. Below, we review representative developments in each direction, identify their limitations, and motivate the architectural choices underlying DREAM.}

\textcolor{black}{
\subsection{Text Modeling for Cross-Modal Retrieval}}
\textcolor{black}{
The use of natural language as a retrieval query emerged in response to the limitations of visual-only systems, such as content-based image retrieval (CBIR), which relied heavily on color, texture, and structural descriptors. Although deep learning significantly improved CBIR by enabling more semantic feature extraction, the visual modality alone often failed to capture user intent when scenes contained ambiguous context or irrelevant background content. As a result, text-guided retrieval methods gained prominence due to the richer and more explicit semantic information provided by natural language.}
\textcolor{black}{
Early text-to-image retrieval (TIR) relied on shallow statistical correlation models, including latent semantic indexing and canonical correlation analysis \cite{chandrika2010multi, shao2017towards, wang2019cross}, which struggled to capture high-level semantics and non-linear relationships between text and images. Deep learning approaches subsequently introduced CNN-RNN pipelines, such as the model proposed by Karpathy and Fei-Fei \cite{karpathy2015deep}, enabling joint embedding learning and improving alignment between modalities. The introduction of transformers and large-scale multimodal pretraining further advanced this direction. In particular, CLIP \cite{radford2021learning} demonstrated that contrastive training on large text-image corpora can yield highly transferable textual embeddings, even though its text encoder is trained without any masked or autoregressive language-modeling objective. BLIP \cite{li2022blip} extended multimodal understanding by incorporating generative supervision that enriches textual semantics, and BridgeTower \cite{xu2023bridgetower} improved cross-modal fusion through deeper cross-layer interactions.
}

\textcolor{black}{
Despite these advancements, most retrieval-oriented text encoders derive their linguistic representations from either masked language modeling, as in BLIP and BridgeTower, or from architectures optimized solely through contrastive objectives, as in CLIP and Long-CLIP \cite{zhang2024long}. Masked modeling captures fine-grained local context but provides limited control over global sentence structure, while contrastive-only encoders optimize for sentence-level alignment but do not explicitly model token-wise dependencies. Other approaches, such as TeachText \cite{croitoru2021teachtext}, enhance semantic diversity through distillation, yet still rely on a single modeling perspective determined by the teacher network. As a result, existing text encoders tend to emphasize either local semantics or global structure, making it difficult to represent queries involving multi-stage activities, compositional phrasing, or subtle semantic distinctions.
}
\textcolor{black}{
These limitations highlight the need for a textual modeling strategy that jointly captures fine-grained dependencies and global semantic organization within a unified framework. DREAM addresses this gap by introducing a dual-path semantic modeling approach that integrates complementary linguistic cues, enabling richer and more discriminative textual representations tailored for alignment with dynamic video content.}
\textcolor{black}{
\subsection{Video Modeling and Efficient Spatiotemporal Attention}}
\textcolor{black}{
Text-to-video retrieval (TVR) introduces the additional challenge of reasoning over motion and temporal structure. Early methods such as Two-Stream CNNs \cite{simonyan2014two} and C3D networks \cite{tran2015learning} learned coarse spatiotemporal features but were constrained by limited receptive fields and difficulty modeling long-term dependencies. Transformer-based architectures led to significant progress by using self-attention to capture temporal relations explicitly. TimeSformer \cite{bertasius2021space}, for example, factorized attention into spatial and temporal components to improve efficiency, while ViT-based video transformers extended image-level architectures to video settings. CLIP4Clip \cite{luo2022clip4clip} further demonstrated that CLIP embeddings could be transferred to video retrieval by aggregating frame-level features, and frameworks such as VILT-CLIP \cite{wang2024vilt} and MV-Adapter \cite{jin2024mv} proposed lightweight fusion modules to enhance cross-modal interactions without excessive computational overhead.}

\textcolor{black}{
Despite these advances, many of these models employ flat global attention or uniform temporal aggregation, treating all tokens equally regardless of their semantic salience. This uniformity can limit a model's ability to capture subtle transitions, multi-stage actions, or object-action interactions, and often leads to high computational cost when processing densely sampled videos. To overcome these issues, recent methods have explored structured or hierarchical mechanisms. MUSE \cite{tang2025muse} introduced multi-scale temporal transformers to model varying temporal resolutions UMP \cite{zhang2024ump} incorporated modality-aware prompts to guide attention and LSDO \cite{zheng2025enhancing} emphasized low-salient object cues to improve fine-grained retrieval. However, these systems typically use fixed attention patterns or single-stage fusion and lack mechanisms for progressive refinement across hierarchical levels.}

\textcolor{black}{The remaining challenge lies in designing video encoders that capture coarse contextual patterns early while refining fine-grained spatial and temporal details in later stages. DREAM addresses this challenge by employing a hierarchical visual encoder equipped with cascaded group attention, allowing for progressive, coarse-to-fine feature refinement. This structure enhances temporal sensitivity and spatial discrimination while maintaining computational efficiency, enabling more robust alignment with complex textual queries. Although prior work has significantly advanced text and video representation learning, important gaps persist. Text encoders often fail to integrate local and global semantics within a single modeling framework, while video encoders frequently rely on flat or weakly structured attention mechanisms that limit their ability to model rich, multi-scale spatiotemporal patterns. DREAM introduces two complementary approaches to address these challenges: a dual-path text encoder that captures both fine-grained and global linguistic semantics, and a hierarchical video encoder that progressively refines representations through cascaded group attention (CGAT). Together, these components form a unified retrieval framework that advances beyond existing designs and addresses longstanding challenges in text-video alignment.}

\section{Method}
\textcolor{black}{In this section, we introduce our overall approach for TVR. DREAM is designed to effectively capture and align information from both textual and visual modalities. On the language side, we enhance the model's ability to understand and represent complex descriptions. For the visual component, we utilize a computationally optimized backbone that processes high-dimensional video data while preserving spatial and temporal fidelity. Additionally, we introduce mechanisms to better integrate spatial and temporal information across video frames. Together, these design choices lead to improved semantic alignment between text and video. To address the two core challenges fine-grained temporal dependencies and nuanced linguistic structures our design choices in DREAM are purposefully aligned with these motivations. The hierarchical vision encoder, equipped with CGAT, models subtle temporal transitions and multi-scale spatiotemporal cues by progressively refining token representations across stages and recursively fusing them through coarse-to-fine attention groups. This enables the model to track object, local transitions and temporal context that are often overlooked in single-stage or static pooling approaches. On the language side, the dual-stream hybrid modeling strategy is introduced to capture both short-range syntactic cues and long-range compositional structure. MLM resolves phrase-level ambiguities by focusing on masked tokens, while PLM models global ordering and semantic relations, enabling the encoder to interpret complex multi-event descriptions. Together, these components ensure that DREAM effectively addresses the temporal and linguistic challenges inherent in text-video retrieval. The following subsections provide detailed insight into each component of our proposed framework.}

\begin{figure*}[t]
    \centering
    \includegraphics[width=\textwidth, height=13 cm]{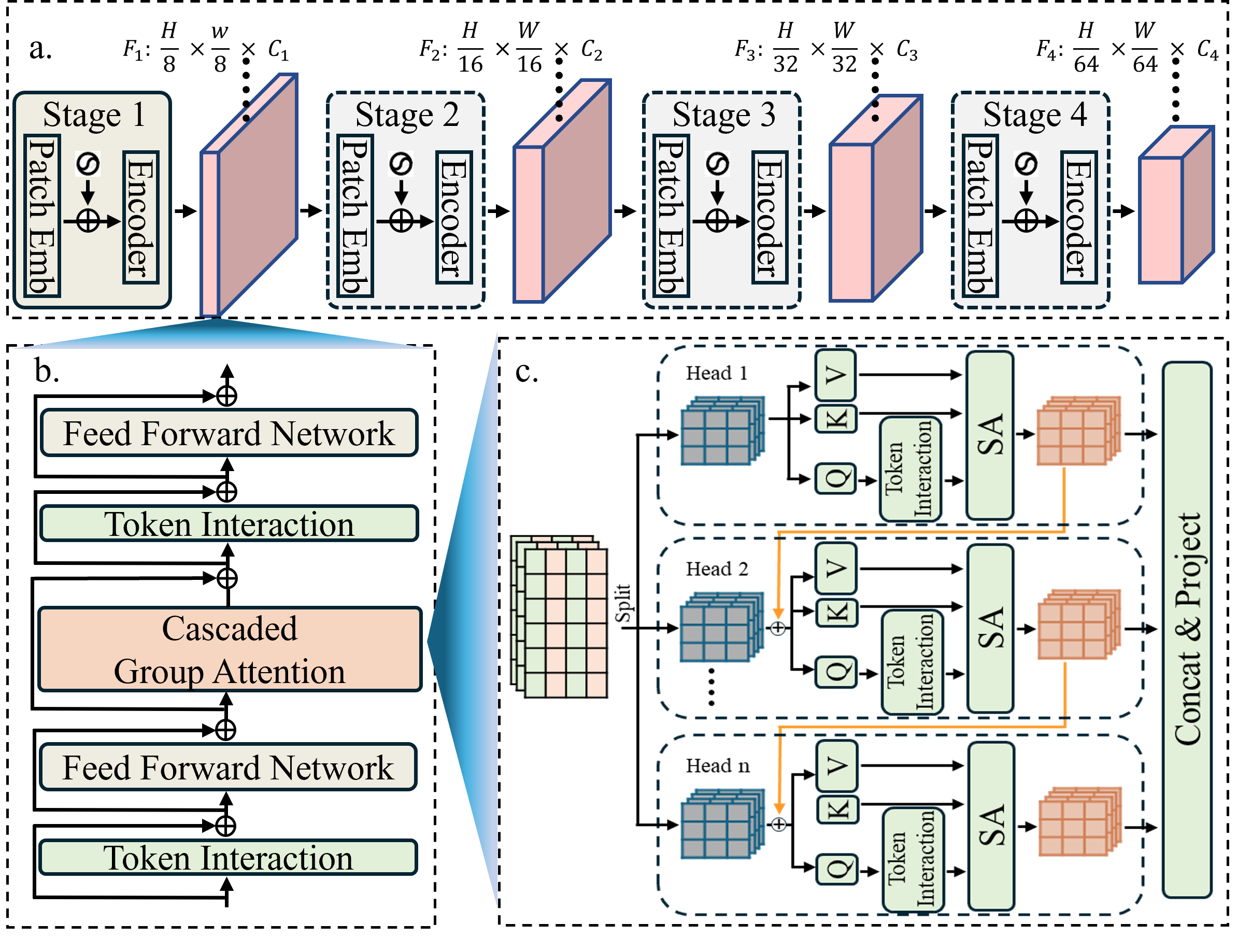}
    \caption{Overview of the visual feature encoder architecture. The top (a) shows a four-stage hierarchical transformer that encodes video frames into multiscale token representations. The bottom-left (b) illustrates the Cascaded Group Attention module, which enhances feature refinement through token interaction and attention. The bottom-right (c) section shows grouped tokens processed across multiple attention heads for feature integration.}
    \label{fig:Vision Encoder}
\end{figure*}
\subsection{Multi-Stage Visual Feature Extraction}
Understanding and modeling visual semantics is fundamental to effective TVR, where the goal is to align high-dimensional visual content with natural language descriptions. To bridge the semantic gap between textual queries and visual content in video retrieval, we design a hierarchical vision encoder that effectively transforms raw frame sequences into rich, discriminative feature embeddings. \textcolor{black}{Specifically, the video frames \( \{ I_t \}_{t=1}^{T} \), where each frame \( I_t \in \mathbb{R}^{H \times W \times 3} \), are processed through a series of stages, generating visual representations denoted as \( Z_s \in \mathbb{R}^{N_s \times d_s} \) at each stage \( s \). Here, \( N_s \) is the number of tokens at stage \( s \), and \( d_s \) is the feature dimension at that stage.} DREAM builds on the strengths of multi-stage transformers, CGAT, and token interaction mechanisms to progressively distill spatiotemporal information into compact visual representations. The architecture is depicted in Fig.~\ref{fig:Vision Encoder}. \par

\begin{figure*}[!h]
    \centering
    \includegraphics[width=\textwidth, height=13 cm]{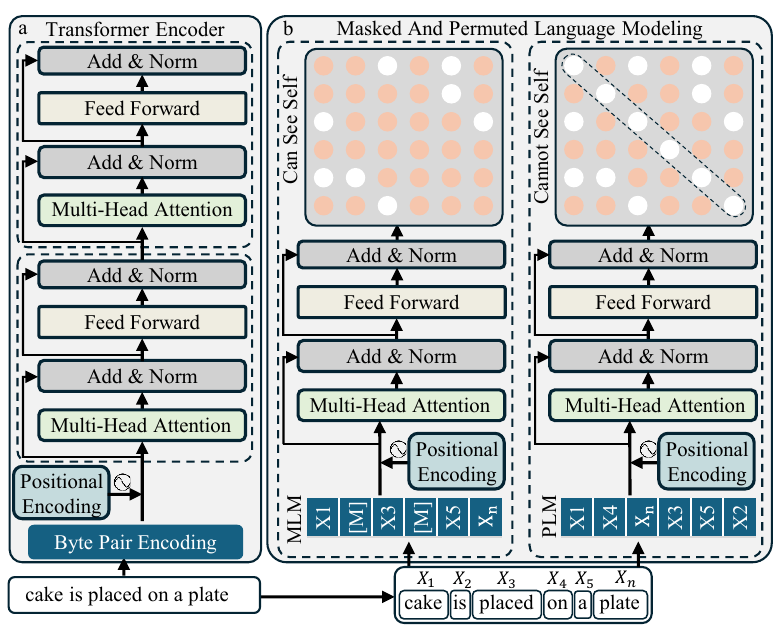}
    \caption{\textcolor{black}{Text encoder architecture illustrating two complementary language modeling strategies within a multi-head attention framework. (a) A transformer-based encoder processes input text, which is first tokenized using byte pair encoding and enriched with positional encodings. (b) The architecture branches into two parallel modeling objectives: masked language modeling (MLM), where selected tokens are masked and predicted based on visible context (including self), and permuted language modeling (PLM), where input tokens are shuffled and processed through a dual-stream attention mechanism. In PLM, the content stream can see all tokens, while the query stream cannot attend to itself.}}
    \label{fig:TextEncoder}
\end{figure*}

The vision encoder operates over a sequence of video frames \( \{I_t\}_{t=1}^{T} \), where each frame \( I_t \in \mathbb{R}^{H \times W \times 3} \) is first partitioned into fixed-size non-overlapping patches. At each stage \( s \in \{1, 2, 3, 4\} \), a patch embedding layer projects the flattened patches into a token embedding space via a linear transformation \( E_s: \mathbb{R}^{p^2 \cdot 3} \rightarrow \mathbb{R}^{d_s} \), where \( p \) is the patch size and \( d_s \) denotes the feature dimension at stage \( s \). \textcolor{black}{The resulting token sequence is then processed by a transformer encoder block \( \mathcal{\vartheta}_s \), producing hidden representations \( Z_s = \mathcal{\vartheta}_s \left( E_s\big(\text{patches}(I_t)\big) \right) \).}
\textcolor{black}{
In practice, the input to stage $s$ is the refined token representation produced by stage $s{-}1$, enabling progressive hierarchical abstraction rather than independent processing of raw frames.
}

\textcolor{black}{To enhance intra- and inter-token communication beyond standard self-attention, 
each stage is augmented with a TokenInteract module. 
Given input tokens $Z \in \mathbb{R}^{N \times d}$, the module performs a lightweight 
local transformation followed by pairwise token interaction. First, a non-linear token-wise projection mixes local channel information using eq. 1:}
\textcolor{black}{\begin{equation}
Z^{\text{TI}} = \sigma(Z W_{1}) W_{2},
\label{eq:ti_local}
\end{equation}
where $W_{1} \in \mathbb{R}^{d \times d_h}$ and $W_{2} \in \mathbb{R}^{d_h \times d}$ 
are learnable matrices, $d_h$ is hidden dimension, and 
$\sigma(\cdot)$ is GELU for non-linearity. Furthermore to capture lightweight pairwise relations, we compute an affinity matrix through feature similarity with eq. 2 which is aggregated for contextual evidence in eq. 3:
\begin{equation}
A = \text{softmax}\!\left( \frac{ZZ^{\top}}{\sqrt{d}} \right),
\label{eq:ti_affinity}
\end{equation}
\begin{equation}
Z^{\text{pair}} = A Z.
\label{eq:ti_pair}
\end{equation}
The output of the TokenInteract module is then formed in eq. 4 by combining the two components:
\begin{equation}
\text{TokenInteract}(Z) = Z^{\text{TI}} + Z^{\text{pair}}.
\label{eq:ti_combine}
\end{equation}
Thus, the overall token interaction step becomes:
\begin{equation}
Z' = Z + \text{TokenInteract}(Z), \quad 
Z'' = Z' + \text{FFN}(Z'),
\label{eq:ti_final}
\end{equation}
where $\text{FFN}(\cdot)$ denotes the position-wise feedforward network. 
This module introduces localized contextualization and lightweight 
pairwise refinement prior to the deeper self-attention computation.}

\textcolor{black}{Central to our encoder design is the CGAT mechanism, which fuses multiscale features from different encoder stages in a progressive fine-to-coarse manner. Feature maps from all stages are partitioned into non-overlapping spatial windows, forming groups \( \{G_k\}_{k=1}^{K} \). Each group corresponds to a fixed spatial region of the frame (e.g., 8×8, 16×16) patches depending on the stage. This spatial grouping ensures computational efficiency and provides a natural hierarchy for progressively refining local details into broader semantic representations.
Unlike conventional visual encoders, the CGAT mechanism explicitly models fine-grained temporal dependencies by enabling hierarchical information flow from lower-level local groups to higher-level semantic groups. This bottom-up refinement allows the encoder to capture micro-temporal variations, object interactions, and action transitions across frames, elements that are often missed by standard CLIP-based encoders or temporally pooled representations. To achieve this, the outputs from lower-level groups are recursively injected into higher-level groups, promoting progressive fusion of local and global semantics. Within each group, standard self-attention (\(\partial\)) is performed using learned queries, keys, and values \( (Q, K, V) \) derived from the grouped tokens, as formalized in Eq.~6.}

\begin{equation}
\partial(Q, K, V) = \text{softmax}\left(\frac{QK^\top}{\sqrt{d}}\right)V
\label{eq:2}
\end{equation}

\textcolor{black}{Each group output is then recursively refined through bottom-up fusion. Specifically, self-attention is first applied within each group to model intra-group dependencies, after which the output is refined using the representation from group \( k{-}1 \), enabling information flow from lower to higher semantic levels.} This recursive formulation is written as in Eq.~7

\begin{equation}
Z_k = \text{SA}(Q_k, K_k, V_k) + f(Z_{k-1})
\label{eq:3}
\end{equation}

where \( f(\cdot) \) denotes a linear projection or residual mapping. Finally, the outputs of all group attention branches are concatenated along the token dimension and passed through a projection layer \( W_p \in \mathbb{R}^{d \times d} \), yielding a unified video embedding presented in eq. 8

\textcolor{black}{\begin{equation}
\mathcal{V}= W_p \cdot [Z_1 \| Z_2 \| \cdots \| Z_K]
\label{eq:4}
\end{equation}}
where \( \| \) denotes token-wise concatenation. This embedding \(\mathcal{V} \in \mathbb{R}^{N' \times d} \), where \( N' \) is the total number of tokens after concatenation, captures multi-scale visual information and serves as the visual representation used for alignment with textual embeddings in downstream retrieval tasks.

\subsection{Dual-Stream Language Modeling}
Capturing the nuanced semantics of natural language is crucial for bridging the gap between textual queries and dynamic visual content in TVR, particularly when addressing linguistic ambiguity in the context of dynamic visual events. To solve this, we design a hybrid text encoder that enhances the representational strength of CLIP's transformer with a dual-language modeling mechanism.
The hybrid MLM–PLM structure is specifically designed to capture both fine-grained local dependencies and global structural semantics. MLM focuses on resolving localized dependencies, such as verb-object relationships, by reconstructing randomly masked tokens, while PLM models longer-range syntactic structures by predicting tokens within a permuted sequence. This dual approach allows the model to handle both short-range linguistic patterns and complex, global dependencies, directly addressing the ambiguity and compositionality often present in natural language queries for video retrieval. 
As illustrated in Fig.~\ref{fig:TextEncoder}(b), this dual-path approach strengthens the encoder’s capacity to understand both local and global linguistic cues, providing a richer semantic representation that aligns well with the complex structure of video content. We formalize the input caption as a tokenized sequence derived from the video’s textual annotation, as shown in eq. 9.

\textcolor{black}{\begin{equation}
X = \{x_1, x_2, \dots, x_T\}, \quad x_i \in \mathcal{W}
\end{equation}}

Where $T$ is the sequence length and  $\mathcal{W}$ denotes the vocabulary. This sequence is passed through the CLIP text encoder, denoted as $\mathcal{E}_{\text{CLIP}}$ in eq. 10 to obtain contextualized embeddings:
\begin{equation}
H_{\text{clip}} = \mathcal{E}_{\text{CLIP}}(X), \quad H_{\text{clip}} \in \mathbb{R}^{T \times d}
\end{equation}

To refine the semantic representation, we apply MLM and PLM objectives in parallel. For MLM, a subset $\mathcal{M}$ of tokens is masked and the model is trained using the remaining visible context as given in eq. 11
\begin{equation}
\mathcal{L}_{\text{MLM}} = - \sum_{i \in \mathcal{M}} \log P(x_i \mid \tilde{X}_{\text{MLM}})
\end{equation}

This encourages the encoder to infer token-level semantics based on surrounding context.
In contrast, in eq. 12 PLM  applies a random permutation $\pi$ to the input sequence indices and trains the model to autoregressively predict the permuted tokens.
\begin{equation}
\mathcal{L}_{\text{PLM}} = - \sum_{t=1}^{T} \log P\left(x_{\pi(t)} \mid x_{\pi(1)}, \dots, x_{\pi(t-1)}\right)
\end{equation}

To effectively model dependencies while preventing information leakage in the PLM path, we incorporate a dual-stream attention mechanism, as illustrated in Fig.~\ref{fig:TextEncoder}(b). In addition to CLIP’s text encoder, our architecture introduces two parallel attention streams: a content stream, which provides keys and values using full context, and a query stream, which restricts tokens from attending to themselves, thereby enforcing a causal structure. For each token $x_{\pi(t)}$ in the permuted sequence, the attention computation follows the standard formulation in Eq. 6
In this setup, the content stream generates $K$ and $V$ from the entire sequence, while the query stream computes $Q$ using a masked view that excludes $x_{\pi(t)}$ itself. This dual-stream design ensures that autoregressive prediction remains faithful to the temporal structure of the permuted sequence. By decoupling the query and context perspectives, the model effectively learns global structure without information leakage, thereby enhancing the semantic richness of the resulting text representations.

Finally, the outputs from the MLM and PLM branches are fused with the original CLIP embeddings using residual summation as shown in Eq.~13:

\begin{equation}
H_{\text{text}} = H_{\text{clip}} + \lambda \cdot (H_{\text{MLM}} + H_{\text{PLM}})
\end{equation}

where $H_{\text{MLM}}$ and $H_{\text{PLM}}$ denote the hidden representations produced by the MLM and PLM branches, respectively, and $\lambda$ is a scalar that balances their contribution. The final embedding $H_{\text{text}} \in \mathbb{R}^{T \times d}$ encodes multimodal priors, local context, and global semantics, making it well-suited for alignment with complex visual content in downstream retrieval tasks. \textcolor{black}{While our PLM branch adopts the dual-stream attention mechanism inspired by XLNet \cite{yang2019xlnet}, our implementation differs in two key ways: (i) it is embedded within CLIP’s transformer encoder rather than a standalone language model, and (ii) it is trained jointly with video embeddings under a retrieval objective. This retrieval-specific conditioning ensures that the learned global dependencies are not generic but tuned to align with temporally dynamic video content. DREAM introduces a hybrid language encoding framework that combines MLM and PLM within a contrastive retrieval setting, a first in video-text retrieval. This approach allows DREAM to address complex challenges, such as multi-event query alignment and fine-grained temporal modeling, which traditional single-path models cannot capture.} The MLM and PLM objectives are used as auxiliary training losses rather than being directly added to the text embeddings. Therefore, the overall training objective is formulated as Eq.~14, where $\mathcal{L}_{\text{ret}}$ denotes the video-text retrieval loss.

\begin{equation}
\mathcal{L}_{\text{total}} =
\mathcal{L}_{\text{ret}}
+  \mathcal{L}_{\text{MLM}}
+ \mathcal{L}_{\text{PLM}}
\end{equation}



\begin{figure*}[t]
    \centering
    \includegraphics[width=\textwidth]{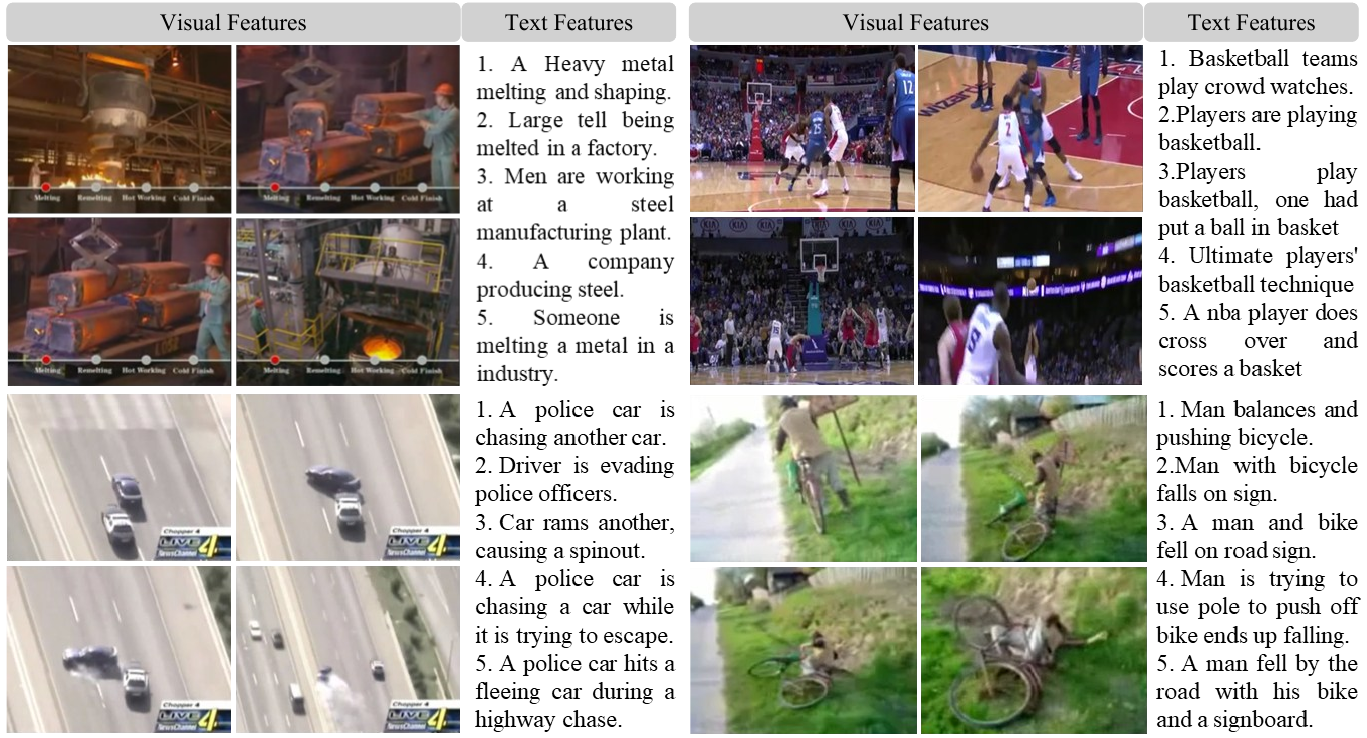}
    \caption{Examples of video-text pairs from benchmark datasets. Each clip is represented by four sampled frames (left) and five corresponding human-annotated sentences (right). The first two examples (Top Row) are drawn from the MSR-VTT dataset, while the remaining examples are from the MSVD dataset.}
    \label{fig:Dataset samples}
\end{figure*}
\subsection{Cross-Modality Alignment}
A fundamental requirement for TVR is the construction of a unified embedding space in which heterogeneous modalities visual and textual data can be directly compared based on their semantic content. This necessitates the projection of both video and language representations into a common latent space wherein cross-modal alignment is both geometrically meaningful. Let $\mathcal{V} \in \mathbb{R}^{d}$ denote the aggregated video embedding and $\mathcal{T} \in \mathbb{R}^{d}$ the corresponding textual embedding, both derived from their respective modality-specific encoders. To quantify the semantic relatedness between these representations, we employ the cosine similarity in Eq.~15, a scale-invariant metric that captures the angular distance between two vectors in high-dimensional space
\textcolor{black}{\begin{equation}
    S(\mathcal{V}, \mathcal{T}) = \frac{\mathcal{V} \cdot \mathcal{T}}{\|\mathcal{V}\| \|\mathcal{T}\|}
\end{equation}}
Following the groundwork established in \cite{luo2022clip4clip} we aggregate frame-level features using mean pooling, which has been empirically validated as an effective and efficient strategy for making video features comparable with textual embeddings. This ensures that both video and textual representations are aligned in a common latent space prior to similarity computation. Furthermore both embeddings undergo $\ell_2$ normalization to constrain them to the unit hypersphere, thereby ensuring that similarity comparisons reflect pure alignment rather than disparities. During inference, this similarity metric is evaluated between a given textual query and the video embeddings within the retrieval corpus. The resulting scores are then ranked to identify the top-$k$ candidate videos whose semantic representations exhibit maximal alignment with the input query.
\textcolor{black}{A key design decision in the DREAM framework is the use of global video-text alignment instead of frame-word or region-phrase correspondence. Methods, such as token-wise cross-attention, prevent efficient pre-indexing of video representations, while Global  design enables real-time retrieval by decoupling video and text embeddings. CGAT mechanism in the DREAM implicitly attends to local spatiotemporal details, allowing the global token to capture essential features for temporally complex queries. Additionally, global-alignment approach is more robust to asynchronous data in real-world datasets, where fine-grained alignment methods struggle with loose temporal synchronization.}

\section{Results and Discussion}
This section provides a comprehensive evaluation of the proposed method using publicly available benchmark datasets for video retrieval. We assess the performance from multiple perspectives, incorporating both quantitative and qualitative analyses. The following subsections offer a detailed discussion of the results obtained through this evaluation.
\subsection{Experimental Setup:}
All the experiments were conducted on a PC running the Ubuntu operating system, equipped with four NVIDIA RTX A5000 GPUs, 128 GB RAM, and an AMD Ryzen 9 3900 24-core processor operating at a base clock speed of 4.2 GHz and 5.3 GHz boost frequency. The proposed framework is implemented in Python 3.10 using PyTorch \cite{paszke2019pytorch} version 2.4+cu118. The training and evaluation procedures including a batch size of 32 for training and 16 for validation, a learning rate of 1e-4, and a total of 5 training epochs. We utilized a maximum of 12 frames per video and 32 tokens per caption following \cite{luo2022clip4clip} . The Adam \cite{kingma2014adam} optimizer was used for training, and inference was performed with frame-based inputs at a rate of 1 frame per second. \textcolor{black}{All experiments were repeated with three random seeds. We report mean ± standard deviation across runs to ensure statistical robustness.}

\begin{table}[t]
\centering
\caption{Ablation study on the MSRVTT dataset. We report R@1 (\%) while incrementally adding key components to the baseline. \ding{51} = component included, \ding{55} = excluded.  MLM and PLM denote Masked and Permuted Language Modeling, respectively, while CGAT stands for Cascaded Group Attention and HVE for the Hierarchical Vision Encoder}
\label{tab:ablation}

\begin{tabular}{lccccc|c}
\hline
\textbf{Exp} & \textbf{MLM} & \textbf{PLM} & \textbf{CLIP Text} & \textbf{CGAT} & \textbf{HVE} & \textbf{R@1} \\
\hline
1            & \ding{55} & \ding{55} & \ding{51} & \ding{55} & \ding{55} & \textcolor{black}{44.2 $\pm$ 0.13}\\
2                           & \ding{51} & \ding{55} & \ding{51} & \ding{55} & \ding{55} & \textcolor{black}{45.5 $\pm$ 0.11} \\
3                           & \ding{55} & \ding{51} & \ding{51} & \ding{55} & \ding{55} & \textcolor{black}{45.2 $\pm$ 0.12} \\
4                    & \ding{51} & \ding{51} & \ding{51} & \ding{55} & \ding{55} & \textcolor{black}{48.1 $\pm$ 0.11} \\
5      & \ding{51} & \ding{51} & \ding{51} & \ding{51} & \ding{55} & \textcolor{black}{48.8 $\pm$ 0.13} \\
\hline
\textbf{6} & \ding{51} & \ding{51} & \ding{51} & \ding{51} & \ding{51} & \textcolor{black}{\textbf{49.4 $\pm$ 0.12}} \\
\hline
\end{tabular}
\end{table}

\subsection{Evaluation Metrics:}
To evaluate the performance of DREAM, we utilize three standard metrics commonly used in information retrieval and ranking task. Recall at K (R@K), Median Rank (Md@R), and Mean Rank (Mn@R). These metrics provide a comprehensive understanding of the effectiveness and efficiency of our approach.

\subsubsection{Recall at K (R@K)}
Recall at K (R@K) measures the proportion of queries for which at least one relevant item appears in the top-\( K \) retrieved results. Let \( N \) denote the total number of queries, and let \( r_i^{(K)} \in \{0, 1\} \) be an indicator variable that equals 1 if a relevant item for query \( i \) is retrieved within the top-\( K \) results, and 0 otherwise. Then, R@K is defined as in Eq.~16

\begin{equation}
\text{R@}K = \frac{1}{N} \sum_{i=1}^{N} r_i^{(K)}
\end{equation}

This metric captures how often the retrieval system succeeds within the top-\( K \) positions, aligning with user expectations in practical search scenarios.

\subsubsection{Median Rank (Md@R)}
The Median Rank (\(Md@R\)) is the median of the ranks of the first relevant item retrieved for all queries. It provides a robust central tendency measure of the ranks and is less sensitive to outliers compared to the mean. A lower \( Md@R \) value indicates that the relevant items are ranked closer to the top on average. This metric is useful for understanding typical query performance without being skewed by extreme cases.

\subsubsection{Mean Rank (Mn@R)}

The Mean Rank (\(Mn@R\)) calculates the average rank of the first relevant item retrieved for all queries. It is calculated as in Eq.~17

\begin{equation}
\text{Mn@R} = \frac{\sum_{i=1}^N r_i}{N}
\end{equation}

where \( r_i \) is the rank of the first relevant item for the \( i \)-th query, and \( N \) is the total number of queries. While \(Mn@R\) is sensitive to outliers, it provides insight into the overall ranking performance across the dataset.

\subsection{Datasets:}
The proposed method was evaluated using three benchmark datasets: MSRVTT \cite{xu2016msr}, MSVD \cite{chen2011collecting} and LSMDC\cite{rohrbach2015long}. Fig.~\ref{fig:Dataset samples} shows sample videos and their annotated sentences. MSRVTT is a large-scale web video dataset comprising 10,000 videos totaling 41.2 hours and 200K clip-sentence pairs. Each video is accompanied by 20 natural language annotations, with an average duration of 30 seconds per video. For evaluation, the test set used is ‘test1k-A’, which contains 1,000 clip-text pairs following the JSFusion setup \cite{yu2018joint}. MSVD contains 1,970 videos, each with a length that ranges from one to 62 seconds. Train, validation and, test splits contain 1,200, 100, and 670 videos, respectively. Each video has approximately 40 associated sentences in English. \textcolor{black}{The LSMDC\cite{rohrbach2015long} dataset consists of 118,081 video clips, each ranging from 2 to 30 seconds in length. These clips are extracted from 202 movies. The validation set includes 7,408 videos, while the test set contains 1,000 videos, all sourced from movies that are separate from those used in the training and validation sets.}

\begin{table}
\centering

\caption{Text-Video retrieval Performance Comparison of Methods on MSRVTT Dataset. Second best is \underline{underlined}.}
\label{tab:MSRVTTT2V}
\resizebox{\columnwidth}{!}{%
\begin{tabular}{l c c c c c c}
\toprule
 \textbf{Methods} & \textbf{Year} & \textbf{R@1~$\uparrow$} & \textbf{R@5~$\uparrow$} & \textbf{R@10~$\uparrow$} & \textbf{MdR~$\downarrow$} & \textbf{MnR~$\downarrow$}  \\ 
\midrule
CT-SAN \cite{yu2017end}               & 2017          & 4.4           & 16.6          & 22.3           & 35.0            & -              \\
JSFusion \cite{Yu_2018_ECCV}          & 2018          & 10.2          & 31.2          & 43.2           & 13.0            & -              \\
CE \cite{liu2019use}                  & 2019          & 20.9          & 48.8          & 62.4           & 6.0             & 28.2           \\
MMT \cite{gabeur2020multi}            & 2020          & 24.6          & 54.0          & 67.1           & 4.0             & -              \\
TT-CE+ \cite{Croitoru_2021_ICCV}      & 2021          & 29.6          & 61.6          & 74.2           & 3.0             & -              \\
Clip4Clip \cite{luo2022clip4clip}     & 2022          & 42.1          & 71.9          & 81.4           & 2.0             & 15.7           \\
ESAT \cite{wu2023text}                & 2023          & -             & 57.9          & 69.9           & 4.0           & 21.8           \\
Zhu et al. \cite{zhu2023complementarity}       & 2023          & 25.5          & 51.7          & 62.6           & 5.0           & -              \\
Align and Tell \cite{wang2022align}   & 2023          & 43.2          & 72.3          & 81.5           & 2.0           & -              \\
Lee et. \cite{lee2024text}            & 2024          & 28.7          & 58.6          & 70.9           & 4.0             & -              \\
UMP \cite{zhang2024ump}               & 2024          & 43.7          & 70.0          & 79.7           & -             & 15.2           \\
LSDO \cite{zheng2025enhancing}        & 2025          & 46.2          & 74.0          & 83.5           & 2.0             & 13.5           \\

CMA \cite{jiang2025cross}        & 2025          & 45.0          & 72.3          & 81.0           & 2.0             & 14.6           \\
HTVR \cite{zhang2025htvr}        & 2025          & \underline{48.0}          & \underline{74.7}          & \underline{83.7}           & 2.0             & \underline{12.9}           \\
VideoAligner \cite{feng2026videoaligner}        & 2026          & 45.9          & 73.2          & 82.5          & -             & 13.1           \\
\hline
\textbf{Ours}    & \textbf{-} & \textcolor{black}{\textbf{49.4 $\pm$ 0.12}} & \textcolor{black}{\textbf{78.4 $\pm$ 0.15}} & \textcolor{black}{\textbf{84.3 $\pm$ 0.10}}  & \textcolor{black}{\textbf{2.0 $\pm$ 0.00}}  & \textcolor{black}{\textbf{12.3 $\pm$ 0.11}}  \\
\bottomrule
\end{tabular}
}
\end{table}

\begin{table}[t]
\centering

\caption{Video-Text retrieval Performance Comparison of Methods on MSRVTT Dataset. Second best is \underline{underlined}.}
\label{tab:MSRVTTV2T}
\resizebox{\columnwidth}{!}{%
\begin{tabular}{l c c c c c c}
\toprule
\textbf{Methods} & \textbf{Year} & \textbf{R@1~$\uparrow$} & \textbf{R@5~$\uparrow$} & \textbf{R@10~$\uparrow$} & \textbf{MdR~$\downarrow$} & \textbf{MnR~$\downarrow$}  \\ 
\midrule
 CE \cite{liu2019use}              & 2019          & 20.6          & 50.3          & 64.0           & 5.3           & 25.1           \\
 MMT \cite{gabeur2020multi}               & 2020          & 24.4          & 56.0          & 67.8           & 4.0             & -              \\
TT-CE+ \cite{Croitoru_2021_ICCV}          & 2021          & 32.1          & 62.7          & 75.0           & 3.0           & -              \\
Clip4Clip \cite{luo2022clip4clip}       & 2022          & 43.1          & 70.5          & 81.2           & 3.0             & -              \\
ESAT \cite{wu2023text}             & 2023          & -             & 57.0          & 70.3           & 4.0           & 19.43          \\
Zhu et al.  \cite{zhu2023complementarity}     & 2023          & 25.3          & 53.2          & 63.4           & 5.0             & -              \\
Align and Tell \cite{wang2022align}   & 2023          & 41.6          & 69.3          & 78.6           & 2.0           & -              \\
UMP   \cite{zhang2024ump}           & 2024          & \underline{44.8}          & 69.9          & 80.6           & -             & 11.0           \\
CMA   \cite{jiang2025cross}           & 2025          & 44.6          & \underline{72.1}          & \underline{83.6}           & 2.0             & \underline{9.8}           \\
\hline
\textbf{Ours}    & \textbf{-} & \textcolor{black}{\textbf{45.7 $\pm$0.12}} & \textcolor{black}{\textbf{72.9 $\pm$0.14}} & \textcolor{black}{\textbf{84.2 $\pm$0.11}}  & \textcolor{black}{\textbf{2.0 $\pm$0.00}}    & \textcolor{black}{\textbf{9.3 $\pm$0.09}}  \\
\bottomrule
\end{tabular}
}
\end{table}

\subsection{Ablation Study:}
We perform the ablation study on the MSRVTT dataset to evaluate the individual contributions of each component in our framework. MSRVTT offers a larger scale, greater diversity, and more complex textual annotations, making it a more challenging and representative benchmark for retrieval performance. Table~\ref{tab:ablation} presents Recall@1 (R@1) scores for various configurations of our model, incrementally adding or removing key components.
\textcolor{black}{Starting from CLIP text embeddings and simple temporal pooling, we observe a performance of 44.2\% R@1. Introducing MLM improves this to 45.5\%, indicating its effectiveness in capturing localized semantic cues. Similarly, PLM contributes 1.0\% gain when used alone. When combined, the hybrid MLM+PLM configuration achieves 48.1\%, confirming the benefit of dual-objective text encoding in modeling both local and global dependencies. These gains confirm that the hybrid MLM+PLM design is not merely a reuse of XLNet-style PLM, but a retrieval-oriented adaptation. On the visual side, the introduction of CGAT alone (without hierarchical stages) further improves performance to 48.8\%. This shows that fine-grained spatial reasoning and coarse-to-fine feature fusion enhance video understanding even without structural depth. Finally, incorporating the full hierarchical visual encoder results in the best overall performance, reaching 49.4\% R@1. This confirms that our visual architecture contributes significantly by capturing multi-scale temporal-spatial features that better align with complex textual queries.}
Overall, the ablation study confirms that both sides of the architecture, hybrid text encoding and hierarchical visual modeling are essential for achieving robust video-text alignment and improving retrieval accuracy.

\begin{table}[t]
\centering
\caption{Text-Video Performance Comparison of Methods on MSVD Datasets. Second Best is \underline{underlined}}
\label{tab:MSVDT2V}
\resizebox{\columnwidth}{!}{%
\begin{tabular}{l c c c c c c}
\toprule
\textbf{Methods} & \textbf{Year} & \textbf{R@1~$\uparrow$} & \textbf{R@5~$\uparrow$} & \textbf{R@10~$\uparrow$} & \textbf{MdR~$\downarrow$} & \textbf{MnR~$\downarrow$}  \\ 
\midrule
VSE\cite{faghri2017vse++}              & 2017          & 12.3          & 30.1          & 42.3           & 14.0            & -              \\
Multi Cues\cite{mithun2018learning}      & 2018          & 20.3          & 47.8          & 61.1           & 6.0             & -              \\
CE\cite{liu2019use}              & 2019          & 19.8          & 49.0           & 63.8          & 6.0   & -         \\
TT-CE+ \cite{Croitoru_2021_ICCV}           & 2021          & 25.4          & 56.9          & 71.3           & 4.0             & -              \\
Clip4Clip\cite{luo2022clip4clip}         & 2022          & 46.2          & 76.1          & 84.6           & 2.0             & 10             \\
QB-Norm\cite{bogolin2022cross}           & 2022          & 47.6          & 77.6          & 86.1           & 2.0             & -              \\
Zhu et al.\cite{zhu2023complementarity}       & 2023          & 25.5          & 51.7          & 62.6           & 5.0           & -              \\
Align and Tell\cite{wang2022align}    & 2023          & 47.1          & 77.0          & 85.6           & 2.0           & -              \\
HTVR \cite{zhang2025htvr}        & 2025          & 47.7          & 75.6          & 85.5           & 2.0             & 9.9           \\
LSDO\cite{zheng2025enhancing}              & 2025          & 48.0          & 78.1          & \underline{86.7}           & 2.0           & 8.8            \\
VideoAligner \cite{feng2026videoaligner}        & 2026          & \underline{48.5}          & \underline{78.2}          & 86.5           & -             & \underline{8.6}           \\ 
\hline
\textbf{Ours}    & \textbf{-} & \textcolor{black}{\textbf{49.7 $\pm$0.15}} & \textcolor{black}{\textbf{79.1 $\pm$0.14}} & \textcolor{black}{\textbf{87.3 $\pm$0.12}}  & \textcolor{black}{\textbf{2.0 $\pm$0.00}}  & \textcolor{black}{\textbf{8.5 $\pm$0.07}}  \\
\bottomrule
\end{tabular}
}
\end{table}

\begin{table}[t]
\centering
\caption{Video-Text Performance Comparison of Methods on MSVD Datasets.}
\label{tab:MSVDV2T}
\resizebox{\columnwidth}{!}{%
\begin{tabular}{l c c c c c c } 
\toprule
\textbf{Methods} & \textbf{Year} & \textbf{R@1~$\uparrow$} & \textbf{R@5~$\uparrow$} & \textbf{R@10~$\uparrow$} & \textbf{MdR~$\downarrow$} & \textbf{MnR~$\downarrow$}  \\ 
\midrule
VSE \cite{faghri2017vse++}             & 2017          & 34.7          & 59.9          & 70.0           & 3.0             & -              \\
 Multi Cues\cite{mithun2018learning}       & 2018          & 31.5          & 51.0          & 61.5           & 5.0           & -              \\
TT-CE+ \cite{Croitoru_2021_ICCV}           & 2021          & 27.1          & 55.3          & 67.1           & 4.0             & -              \\
Clip4Clip\cite{luo2022clip4clip}         & 2022          & 56.6          & 79.7          & 84.3           & 1.0             & 7.6            \\
Align and Tell\cite{wang2022align}    & 2023          & 61.8          & 87.5          & \textbf{92.7}           & 1.0           & -              \\
\hline
\textbf{Ours}    & \textbf{-} & \textcolor{black}{\textbf{62.6 $\pm$0.14}} & \textcolor{black}{\textbf{87.9 $\pm$0.12}} & \textcolor{black}{92.5 $\pm$0.10}  & \textcolor{black}{{1.0 $\pm$0.00}}    & \textcolor{black}{\textbf{4.1 $\pm$0.06}}   \\
\bottomrule
\end{tabular}
}
\end{table}

\begin{table}[t]
\centering
{\color{black}
\caption{Text-Video Performance Comparison of Methods on LSMDC Datasets.}
\label{tab:LSMDCT2V}
\resizebox{\columnwidth}{!}{%
\begin{tabular}{l c c c c c c}
\toprule
\textbf{Methods} & \textbf{Year} & \textbf{R@1~$\uparrow$} & \textbf{R@5~$\uparrow$} & \textbf{R@10~$\uparrow$} & \textbf{MdR~$\downarrow$} & \textbf{MnR~$\downarrow$}  \\ 
\midrule
CT-SAN\cite{yu2017end}             & 2017          & 5.1          & 16.3          & 25.2           & 46.0            & -              \\
JSFusion\cite{Yu_2018_ECCV}     & 2018          & 9.1          & 21.2          & 34.1        & 36.0             & -              \\
CE\cite{liu2019use}              & 2019          & 11.2          & 26.9         & 34.8           & 25.3          & -\\
MMT\cite{gabeur2020multi}            & 2020          & 12.9         & 29.9         & 40.1           & 19.3          & 75.0              \\
TT-CE+ \cite{Croitoru_2021_ICCV}           & 2021          & 17.2          & 36.5         & 46.3           & 13.7             & -              \\
Clip4Clip\cite{luo2022clip4clip}         & 2022          & 22.6          & 41.0          & 49.1           & 11.0            & 61.0            \\
Align and Tell\cite{wang2022align}    & 2023          & 23.1          & 41.2          & 49.6           & 11.0          & -              \\
ESAT\cite{wu2023text}        & 2023          & -          & 32.3         & 40.8           & 17.0           & 70.92              \\
UMP\cite{zhang2024ump}              & 2024          & 21.6          & 41.1          & 49.9           & -           & 59.5            \\
LSDO\cite{zheng2025enhancing}              & 2025          & 26.4          & \textbf{44.8}          & 55.1           & 8.0           & 52.1            \\ 
\hline
\textbf{Ours}    & \textbf{-} & \textcolor{black}{\textbf{27.3 $\pm$0.11}} & \textcolor{black}{44.6 $\pm$0.13} & \textcolor{black}{\textbf{56.4 $\pm$0.12}}  & \textcolor{black}{\textbf{8.0 $\pm$0.00}}  & \textcolor{black}{\textbf{49.6 $\pm$0.17}}  \\
\bottomrule
\end{tabular}
}}
\end{table}


\begin{table}
\centering
{\color{black}
 \caption{Video-to-Text Retrieval Performance Comparison on the LSMDC Dataset.}
\label{tab:LSMDCV2T}

\resizebox{\columnwidth}{!}{%
\begin{tabular}{l c c c c c c}
\toprule
\textbf{Methods} & \textbf{Year} 
& \textbf{R@1~$\uparrow$} 
& \textbf{R@5~$\uparrow$} 
& \textbf{R@10~$\uparrow$} 
& \textbf{MdR~$\downarrow$} 
& \textbf{MnR~$\downarrow$} \\
\midrule
CLIP4Clip\cite{luo2022clip4clip} 
& 2022 & 20.6 & 39.4 & 47.5 & 13.0 & 56.7 \\

Align and Tell\cite{wang2022align} 
& 2023 & 21.3 & 40.2 & 50.1 & 11.0 & - \\

ESAT\cite{wu2023text}  
& 2023 & - & 31.9 & 42.2 & 18.0 & 71.61 \\

UMP\cite{zhang2024ump} 
& 2024 & 22.5 & 40.7 & 50.8 & - & 51.7 \\

\midrule
\textbf{Ours} 
& \textbf{-} 
& \textbf{23.1 $\pm$ 0.4} 
& \textbf{40.9 $\pm$ 0.5} 
& \textbf{51.3 $\pm$ 0.6} 
& \textbf{10.0 $\pm$ 0.0} 
& \textbf{51.0 $\pm$ 0.7} \\
\bottomrule
\end{tabular}
}
}
\end{table}

\subsection{SOTA Comparison}
To comprehensively evaluate the effectiveness and generalizability of DREAM, we conduct extensive comparisons against recent state-of-the-art methods on standard video-text retrieval benchmarks. Our evaluation covers both text-to-video and video-to-text retrieval tasks across three datasets: MSRVTT, MSVD and LSMDC. These benchmarks differ in scale and complexity, with MSRVTT offering broader diversity and richer annotations, while MSVD provides more concise descriptions and a smaller but well-curated set of video clips. By benchmarking on both datasets, we aim to demonstrate the robustness of our model under varying retrieval scenarios.

\textbf{MSRVTT}: The results on MSRVTT are presented in Table~\ref{tab:MSRVTTT2V}. Our method consistently outperforms all previous SOTA approaches. For the text-to-video retrieval task, our model obtains an R@1 score of 49.4\%, surpassing the best previously reported result of 46.2\% by LSDO\cite{zheng2025enhancing}. It also achieves strong performance at R@5 (78.4\%) and R@10 (84.3\%), closely matching the best-performing baselines. The Median Rank (MdR) of 2 and a Mean Rank (MnR) of 12.3 further demonstrate the model’s ability to consistently retrieve relevant results early in the ranking list. These improvements can be attributed to our proposed CGAT mechanism, which refines spatial-temporal dependencies in the visual modality and enables the model to construct richer video representations. In contrast to prior approaches, such as CLIP4Clip\cite{luo2022clip4clip} and UMP\cite{zhang2024ump}, which rely on static pooling or basic temporal modeling, our approach utilizes the hierarchical structure of video content to generate more discriminative embeddings.

In the video-to-text retrieval task on MSRVTT, our method in Table~\ref{tab:MSRVTTV2T} again establishes a new performance benchmark, achieving an R@1 score of 45.7\%, R@5 of 72.9\%, and R@10 of 84.2\%. These results outperform previous baselines by a significant margin. The model also maintains an MdR of 2 and reduces MnR to 9.3, the best reported on this dataset. These gains stem from the design of our hybrid language encoder, which combines MLM and PLM in a dual-stream attention setup. This structure allows the model to capture both localized semantic dependencies and global sentence structure, resulting in text embeddings that are more robust to ambiguity and better aligned with temporally structured video content.

\begin{figure*}[b]
    \centering
    \includegraphics[width=1\linewidth]{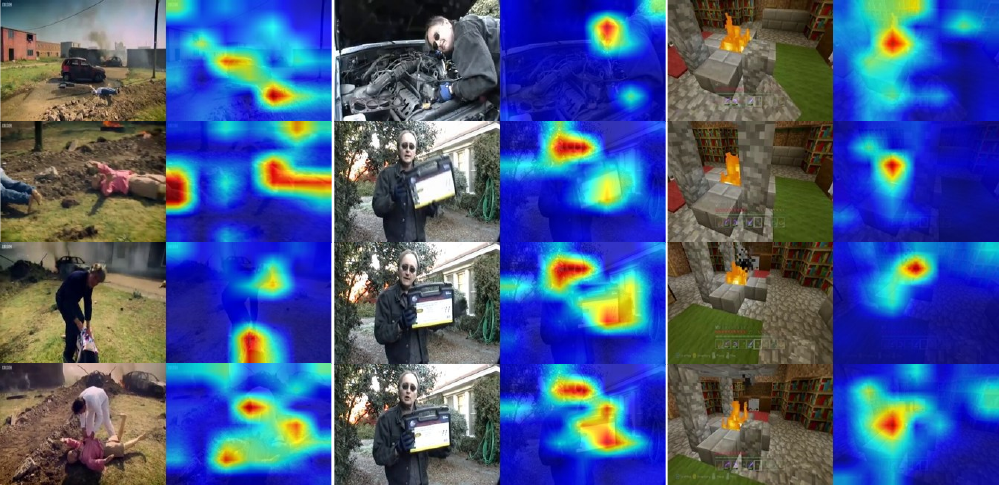}
    \caption{Attention heatmaps from the final transformer layer, illustrating the model’s focus on semantically relevant regions. The left set of examples shows the model attending to key actions in a simulated emergency scenario involving human interaction with a mannequin, capturing motion and contact points. The right set highlights consistent attention on a glowing fireplace within a virtual Minecraft environment, despite variations in viewpoint. These visualizations demonstrate the effectiveness of the CGAT mechanism in localizing salient content, maintaining coherent focus across  scene.}
    \label{fig:heatmap}
\end{figure*}

\textbf{MSVD:} To further validate the generalizability of our proposed framework across datasets of varying scale and complexity, we conduct additional evaluations on the MSVD benchmark. Compared to MSRVTT, MSVD contains fewer training text-video pairs, making retrieval performance more sensitive to model generalization. As shown in Table~\ref{tab:MSVDT2V}, in the TVR setting, our model achieves an R@1 of 49.7\%, R@5 of 79.1\%, and R@10 of 87.3\%. While LSDO~\cite{zheng2025enhancing} performs slightly lower at R@1 (48.0\%), both methods achieve comparable MdR (2.0) and MnR (8.5). These results suggest that our model maintains robust performance even under data-constrained scenarios, and the small dataset size may exaggerate performance differences due to subtle variations in visual or linguistic representations.

In Table~\ref{tab:MSVDV2T}, for the video-to-text retrieval task, our method achieves the best overall performance, with an R@1 of 62.6\%, R@5 of 87.9\%, and R@10 of 92.5\%. While Align and Tell~\cite{wang2022align} slightly outperforms our method on R@10 (92.7\%), our model achieves a lower MnR of 4.1, highlighting its ability to retrieve relevant captions more consistently and with fewer errors. These improvements can be attributed to the synergy between our hierarchical visual encoder and hybrid text modeling approach, which together enhance semantic alignment across modalities.

\textcolor{black}{\textbf{LSMDC:} We also evaluated our method on the LSMDC dataset, with results shown in Table~\ref{tab:LSMDCT2V} for text-to-video retrieval and Table~\ref{tab:LSMDCV2T} for video-to-text retrieval. In the text-to-video task, our model achieves an R@1 score of 27.3\%, outperforming LSDO (26.4\%) and other recent baselines. We also report R@5 (44.6\%) and R@10 (56.4\%), demonstrating consistent performance across different retrieval ranks. In video-to-text retrieval, our model achieves R@1 of 23.1\%, surpassing previous approaches such as CLIP4Clip (20.6\%) and Align and Tell (21.3\%), with strong performance at higher ranks (R@5: 40.9\%, R@10: 51.3\%). These results highlight the effectiveness of our dual-objective text encoder and hierarchical visual encoder in improving retrieval performance across both tasks on the LSMDC dataset.}
\begin{figure*}
    \centering
    \includegraphics[width=\linewidth]{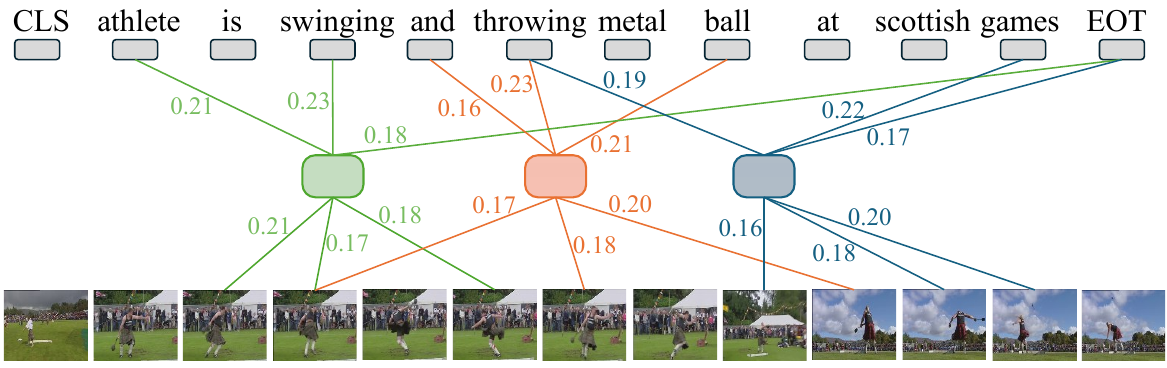}
    \caption{Visualization of attention weights in the alignment module. We select three query embeddings and top-3 attention weights for clearer interpretation. The visualization illustrates how the model captures fine-grained temporal alignment between textual concepts and visual content.}
    \label{fig:attentionheads}
\end{figure*}
\begin{figure*}
    \centering
    \includegraphics[width=\linewidth]{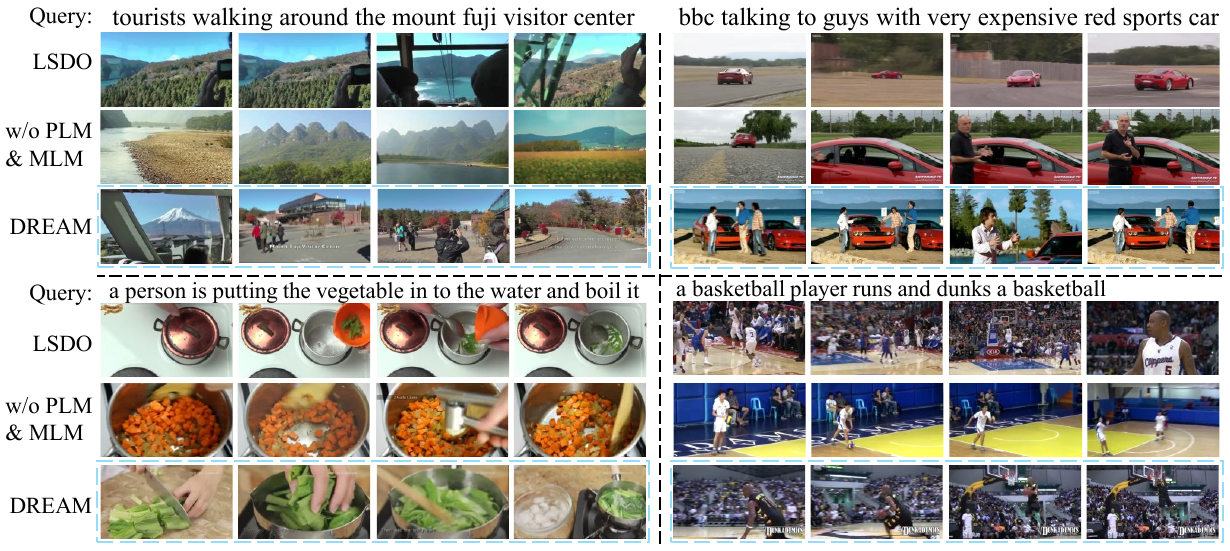}
    \caption{\textcolor{black}{Qualitative comparison of text-to-video retrieval results on the MSRVTT \cite{xu2016msr} dataset. Given each text query, we show the R@1 retrieved video produced by LSDO\cite{zheng2025enhancing}, a variant without MLM and PLM and the proposed DREAM model. Blue box highlights correct results.}}
    \label{fig:qualitative_comparison}
\end{figure*}
\subsection{Qualitative Analysis:}
To better understand how our proposed framework interprets visual information and aligns it with textual semantics, we conduct a qualitative analysis using attention heatmaps and token-frame alignment visualizations on samples from the MSRVTT and MSVD datasets. Fig.~\ref{fig:heatmap} presents attended regions across several video frames, each paired with its corresponding attention activation map. The visualizations demonstrate that the CGAT mechanism consistently highlights semantically meaningful regions relevant to the textual query. For example, in the left sequence, the model successfully identifies and tracks a simulated emergency scenario involving a mannequin and human actors. Attention consistently focuses on critical regions such as the body, limbs, and points of interaction during actions like dragging or assisting, capturing both motion and human-object engagement. In contrast, the right sequence, taken from a Minecraft environment, shows the model's attention remaining tightly centered on a glowing fireplace across varying camera angles and frames. This indicates the model’s robust ability to maintain spatial and temporal coherence, a key strength for aligning dynamic visual content with natural language queries.

To further examine how our model performs fine-grained cross-modal alignment, we present an additional visualization in Fig.~\ref{fig:attentionheads}, which shows the sentence “athlete is swinging and throwing a metal ball at the Scottish Games.” This example demonstrates the model’s capacity to decompose complex textual input into semantically meaningful components that align with temporally distributed video frames. The green head focuses on early-stage cues which correspond to the preparatory phase of the action. The orange head emphasizes the core motion “throwing” linked to the middle frames capturing the throw. The blue head attends to the outcome and context, which align with the final frames showing the airborne object and the event setting. The attention distribution across heads illustrates the model’s ability to distinguish between subject, action, and context within a video, and to associate each component of the query with the most relevant temporal segment.

\textcolor{black}{
In addition to attention visualizations, we performed qualitative comparison across semantic complexity, Fig.~\ref{fig:qualitative_comparison} presents a qualitative comparison of R@1 retrieval results between LSDO, a variant without MLM and PLM, and the proposed DREAM model across scenarios of increasing semantic complexity. The top row (Single Action) corresponds to relatively simple queries involving a single dominant activity, where all methods retrieve visually related videos, however, DREAM exhibits more consistent alignment with both the action and scene context, while baseline methods occasionally emphasize background content missing the tourists. The second row (left) illustrates Human-Object Interaction, where accurate retrieval requires understanding both the manipulated object and the associated action. In this case, LSDO retrieves videos containing relevant objects but fails to capture the precise interaction (e.g., boiling vegetables), whereas DREAM successfully retrieves videos reflecting both object presence and the intended manipulation. The second row (right) corresponds to a Multi-Actor Scene, which is the most challenging scenario due to interactions among multiple agents and temporally ordered actions. For the query ``a basketball player runs and dunks a basketball,'' LSDO retrieves a video where the dunk is performed without the running action, missing a key semantic component. The variant without MLM and PLM retrieves a running player however, the retrieved sport is volleyball rather than basketball, leading to semantic mismatch. In contrast, DREAM correctly retrieves a video capturing both the running motion and dunk action within the appropriate basketball context.}
\begin{figure*}
    \centering
    \includegraphics[width=\linewidth]{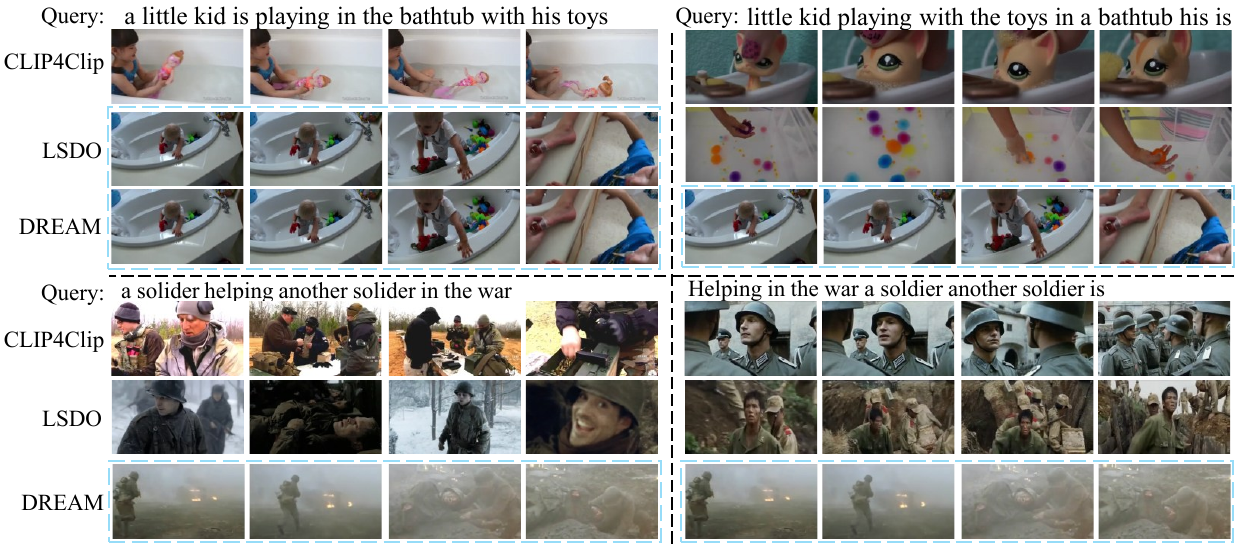}
    \caption{\textcolor{black}{Qualitative comparison of text-to-video retrieval results on the MSRVTT\cite{xu2016msr} dataset. Given each text query, we show the R@1 retrieved video produced by CLIP4Clip\cite{luo2022clip4clip}, LSDO\cite{zheng2025enhancing}, and DREAM model. The left column presents the retrieval results with the original queries, while the right column shows the retrieval results after perturbing the input queries to observe the effect of permutation modeling. Blue box highlights correct results.}}
    \label{fig:qualitative_comparison_permuted}
\end{figure*}

\textcolor{black}{Further to this, Fig.~\ref{fig:qualitative_comparison_permuted} also presents a comparison of retrieval results between the original input query (left column) and the permuted query (right column). Despite changes in word order, DREAM consistently retrieves relevant videos, highlighting its robustness to query perturbations. In contrast, the baseline models and the variant without MLM and PLM objectives show noticeable performance degradation when the query is permuted, emphasizing the role of DREAM's dual-objective modeling in maintaining semantic alignment despite linguistic variations.}

Both visualizations demonstrate that our model’s hierarchical attention design and dual-modality encoding enable it to learn semantically grounded, temporally aware alignments that surpass the flat or diffuse attention patterns typical of prior approaches. Compared to conventional vision transformers or CLIP-based retrieval models, our method achieves sharper, more discriminative attention, reinforcing the superior alignment performance observed in our quantitative evaluations. While most attention patterns are interpretable and consistent, we note that in rare cases such as cluttered or fast-moving scenes attention may become more diffuse. These instances suggest directions for future work, including the integration of motion-specific priors and optical flow cues to further improve temporal reasoning and retrieval robustness.

\section{Conclusion}
\textcolor{black}{In this paper, we propose a novel framework for TVR that comprehensively tackles the semantic and temporal complexity of video content. Our model leverages a hybrid language encoder, combining masked and permuted language modeling to enrich textual representations, alongside a hierarchical vision encoder equipped with CGAT for fine-grained spatial-temporal feature extraction.} This dual-modality design enables robust alignment between natural language queries and dynamic video sequences. Extensive experiments on the MSR-VTT, MSVD and LSMDC benchmarks demonstrate that our method outperforms previous state-of-the-art approaches in both retrieval accuracy and interpretability. We further validated the effectiveness of our approach through qualitative analyses, including heatmap visualizations and token-frame attention alignments, which reveal that our model can consistently localize semantically meaningful content across time. These insights confirm the model’s ability to track complex actions and align them with linguistic structures, making it particularly suitable for real-world multimedia retrieval scenarios. Despite its strong performance, our model has certain limitations. In highly cluttered scenes or under poor lighting conditions, attention maps occasionally become diffuse, reducing alignment precision. Additionally, while the current framework captures temporal context using frame-level modeling, it does not explicitly incorporate motion features such as optical flow or velocity cues, which may limit its performance on fast-paced or subtle temporal events.
\textcolor{black}{
Future work can explore the integration of motion-aware priors and temporal dynamics through video-specific modules, such as 3D convolutions and transformer-based temporal modeling. Furthermore, extending the model to handle multilingual queries, longer video sequences and exploring concept-level alignment and concept permutation/masking could improve the model’s ability to handle more abstract, high-level semantic relationships within videos, allowing for greater flexibility and robustness in text-video retrieval tasks.}

\bibliographystyle{IEEEtran}
\bibliography{references}
\end{document}